\documentclass[11pt,a4paper]{article}
\usepackage[hyperref]{emnlp-ijcnlp-2019}
\usepackage{times}
\usepackage{latexsym}
\usepackage{amsmath}
\usepackage{amsfonts}
\usepackage{bbm}
\usepackage{booktabs}
\usepackage{graphicx}
\usepackage{multirow}
\usepackage{url}
\usepackage{todonotes}

\newcommand\blfootnote[1]{%
  \begingroup
  \renewcommand\thefootnote{}\footnote{#1}%
  \addtocounter{footnote}{-1}%
  \endgroup
}

\aclfinalcopy 

\newcommand{\CM}[2][]{\textcolor{black}{#2}}
\newcommand{\FB}[2][]{\textcolor{black}{#2}}
\newcommand{\OS}[2][]{\textcolor{black}{#2}}

\title{Better Rewards Yield Better Summaries: \\
Learning to Summarise Without References }

\author{
Florian B{\" o}hm$^1$,  
Yang Gao$^{1*}$,  Christian M. Meyer$^1$, \\
\textbf{Ori Shapira$^2$, Ido Dagan$^2$,
and Iryna Gurevych$^1$} \\
$^1$Ubiquitous Knowledge Processing Lab, 
Technische Universit{\" a}t Darmstadt, Germany \\
$^2$Computer Science Department, Bar-Ilan University, Ramat-Gan, Israel \\
{\tt https://www.ukp.tu-darmstadt.de/} \\
{\tt yang.gao@rhul.ac.uk, obspp18@gmail.com, dagan@cs.biu.ac.il}
  \\}

\date{}

\begin{document}
\maketitle
\begin{abstract}
\emph{Reinforcement Learning (RL)} based document summarisation
systems yield state-of-the-art performance in terms 
of ROUGE scores, because 
they directly use ROUGE as the \emph{rewards} during training.
However, summaries with high ROUGE scores often receive
low human judgement.
To find a better reward function that can
guide RL to generate human-appealing  
summaries, 
we learn a reward function from human ratings on 2,500 summaries. 
%
%
Our reward function
only takes the document and system summary as input.
Hence, once trained, it can be used to train RL-based summarisation
systems without using any reference summaries.
We show that our learned rewards have
significantly higher correlation
with human ratings than previous approaches. 
Human evaluation experiments show that,
compared to the state-of-the-art supervised-learning systems
and ROUGE-as-rewards RL summarisation systems,
the RL systems using our learned rewards during training
generate summaries with higher human ratings.
The learned reward function and our
source code 
are available at 
\url{https://github.com/yg211/summary-reward-no-reference}.

\end{abstract}

\section{Introduction}
\label{sec:introduction}
\emph{Document summarisation} aims at generating 
a summary for a long document or multiple documents on the same topic.
\blfootnote{* Since June 2019, Yang Gao is affiliated with Dept. of Computer Science, Royal Holloway, University of London.}
%
%
%
\emph{Reinforcement Learning (RL)} becomes an
increasingly popular technique to build document
summarisation systems in recent years
\cite{DBLP:conf/acl/BansalC18,DBLP:conf/naacl/NarayanCL18,DBLP:conf/emnlp/DongSCHC18}.
Compared to the 
supervised learning paradigm,
which ``pushes'' the summariser to reproduce the reference
summaries at training time, RL directly optimises the
summariser to maximise the \emph{rewards},
which measure the quality of the generated summaries.
%

The \emph{ROUGE} metrics \cite{lin2004rouge} are the 
most widely used rewards in training 
RL-based summarisation systems.
%
ROUGE measures the quality of a generated summary by
counting how many n-grams 
in the reference summaries appear in the generated summary.
ROUGE correlates well with human judgements at 
\emph{system level} \cite{lin2004looking,DBLP:journals/coling/LouisN13},
i.e.\ by aggregating system summaries' ROUGE scores
across multiple input documents, we can reliably 
rank summarisation systems by their quality.
However, ROUGE performs poorly at \emph{summary level}:
given multiple summaries for the same input document,
ROUGE can hardly distinguish the ``good'' summaries from  
the ``mediocre'' and ``bad'' ones
\cite{DBLP:conf/emnlp/NovikovaDCR17}.
Because existing RL-based summarisation systems rely on
summary-level ROUGE scores to guide the optimisation direction,
the poor performance of ROUGE at summary level
severely misleads the RL agents. 
The reliability of ROUGE as RL reward is further
challenged by the fact that most large-scale
summarisation datasets only have one reference summary
available for each input document (e.g. CNN/DailyMail
\cite{DBLP:conf/nips/HermannKGEKSB15,DBLP:conf/acl/SeeLM17}
and NewsRooms \cite{DBLP:conf/naacl/GruskyNA18}).

In order to find better rewards that can guide
RL-based summarisers to generate more human-appealing summaries,
we learn a reward function directly from
human ratings. We use the dataset compiled 
by \citet{chaganty-mussmann-liang:2018:Long}, 
which includes human ratings on 2,500 summaries for 500
news articles from CNN/DailyMail. 
Unlike ROUGE that requires one or multiple reference
summaries to compute the scores,
our reward function only takes the document and the generated
summary as input. 
Hence, once trained, our reward can be used to train
RL-based summarisation systems without any reference summaries.

The contributions of this work are threefold:
\textbf{(i)} We study the summary-level correlation between
ROUGE and human judgement on 2,500 summaries (\S\ref{sec:motivation}), 
explicitly showing that ROUGE can hardly identify the 
human-appealing summaries.
%
\textbf{(ii)} We formulate the reward learning problem
as either a regression or a preference learning problem
(\S\ref{sec:formulation}), and 
explore multiple text encoders and neural architectures 
to build the reward learning model (\S\ref{sec:model}).
\textbf{(iii)}
We show that our learned reward correlates significantly better with
human judgements than ROUGE (\S\ref{sec:reward_quality}),
and that using the learned reward can guide both extractive
and abstractive RL-based summarisers to generate summaries
with significantly higher human ratings than 
the state-of-the-art systems
(\S\ref{sec:summary_evaluation}).
%

\section{Related Work}
\label{sec:related_work}

\paragraph{RL-based summarisation.}
Most existing RL-based summarisers fall into two categories:
\emph{cross-input} systems and \emph{input-specific} systems
\cite{DBLP:conf/ijcai/0023MMG19}.
Cross-input systems learn a summarisation policy
at training time by letting the RL agent interact
with a ROUGE-based reward function. 
At test time, the learned policy is used to generate a summary
for each input document.
Most RL-based summarisers fall into this category
\cite{DBLP:conf/acl/BansalC18,DBLP:conf/naacl/NarayanCL18,DBLP:conf/emnlp/DongSCHC18,DBLP:conf/emnlp/KryscinskiPXS18,DBLP:conf/naacl/PasunuruB18,DBLP:conf/iclr/PaulusXS18}.
As an alternative, input-specific RL 
\cite{rioux2014emnlp,ryang2012emnlp}
does not require reference summaries:
for each input document at test time,
a summarisation policy is trained specifically for
the input, by letting the RL summariser
interact with a heuristic-based reward function,
e.g. ROUGE between the generated summary and the input
document (without using any reference summaries). 
However, the performance of input-specific RL falls 
far behind the cross-input counterparts.

In \S\ref{sec:summary_evaluation} we use our learned
reward to train both cross-input and input-specific RL systems.
A similar idea has been explored by  
\citet{DBLP:conf/ijcai/0023MMG19}, 
but unlike their work that learns the reward from ROUGE scores,
we learn our reward directly from human ratings.
Human evaluation experiments suggest that 
our reward can guide both kinds of RL-based systems to generate
human-appealing summaries without using reference summaries.

The reward learning idea is also related to 
\emph{inverse RL} (IRL) \cite{DBLP:conf/icml/NgR00}. 
By observing some (near-)optimal sequences of actions,
IRL algorithms learn a reward function that is consistent
with the observed sequences. 
In the case of summarisation, human-written reference summaries
are the (near-)optimal sequences, which are expensive to provide.
Our method only needs human ratings on some generated summaries,
also known as the \emph{bandit feedback} \cite{DBLP:conf/acl/KreutzerSR17},
to learn the reward function.
%
Furthermore, when employing certain loss functions 
(see \S\ref{sec:formulation} and Eq.~\eqref{eq:pref_loss}),
our method can even learn the reward function
from preferences over generated summaries, 
an even cheaper feedback to elicit \cite{kreutzer2018b,DBLP:conf/emnlp/GaoMG18}.

\paragraph{Heuristic-based rewards.}
Prior work proposed heuristic-based reward functions
to train cross-input RL summarisers, in order to
strengthen certain properties of the generated summaries.
\citet{DBLP:journals/corr/abs-1904-02321}
propose four reward functions to train 
an RL-based extractive summariser,
including the \emph{question-answering competency} rewards,
which encourage the RL agent to generate
summaries that can answer cloze-style questions.
Such questions are automatically created by 
removing some words in the reference summaries.
Experiments suggest that
human subjects can answer the questions with
high accuracy by reading their generated summaries;
but the human judgement scores of their summaries
are not higher than the summaries generated by
the state-of-the-art supervised system. 
\citet{DBLP:conf/emnlp/KryscinskiPXS18} propose
a simple heuristic that encourages
the RL-based abstractive summariser to generate
summaries with more \emph{novel} tokens,
i.e.\ tokens that do not appear in the input
document. However, both ROUGE and human evaluation scores
of their system are lower than the state-of-the-art 
summarisation systems (e.g.\  \citealp{DBLP:conf/acl/SeeLM17}).
In addition, the above rewards require reference summaries,
unlike our reward that only takes a document
and a generated summary as input.

\paragraph{Rewards learned with extra data.}
\citet{DBLP:conf/naacl/PasunuruB18} propose two novel rewards
for training RL-based abstractive summarisers:
\emph{RougeSal}, which up-weights the salient 
phrases and words detected via a keyphrase classifier,
and \emph{Entail} reward, which gives high 
scores to logically-entailed summaries 
using an entailment classifier.
RougeSal is trained with the SQuAD reading comprehension dataset
\cite{DBLP:conf/emnlp/RajpurkarZLL16},
and Entail is trained with the SNLI \cite{DBLP:conf/emnlp/BowmanAPM15} 
and Multi-NLI \cite{DBLP:conf/naacl/WilliamsNB18} datasets.
Although their system achieves new state-of-the-art
results in terms of ROUGE, it remains unclear whether
their system generates more human-appealing summaries
as they do not perform human evaluation experiments.
Additionally, both rewards require reference summaries.

\citet{DBLP:journals/coling/LouisN13},
\citet{DBLP:conf/emnlp/PeyrardBG17} and \citet{DBLP:conf/naacl/PeyrardG18} build feature-rich 
regression models to learn a summary evaluation 
metric directly from the human judgement scores 
(Pyramid and Responsiveness) 
provided in the TAC'08 and '09 datasets\footnote{\url{https://tac.nist.gov/data/}}.
Some features they use require reference summaries
(e.g.\ ROUGE metrics); the others 
are heuristic-based
and do not use reference summaries
(e.g.\ the Jensen-Shannon divergence between the word 
distributions in the summary and the documents).
Their experiments suggest that with only non-reference-summary-based
features, the correlation
of their learned metric with human judgements is 
lower than ROUGE; with reference-summary-based features,
the learned metric marginally outperforms ROUGE.
In \S\ref{sec:reward_quality}, we show that our reward
model does not use reference summaries but outperforms
the feature-based baseline by \citet{DBLP:conf/naacl/PeyrardG18}
as well as ROUGE. 

Unlike the above work that attempts to
learn a summary evaluation metric, 
the target of our work is to learn a good \emph{reward},
which is not necessarily a good \emph{evaluation metric}.
A good evaluation metric should be able to correctly rank
summaries of different quality levels, while a good reward function
focuses more on distinguishing the best summaries
from the mediocre and bad summaries.
Also, an evaluation metric should be able to evaluate
summaries of different types (e.g. extractive
and abstractive) and from different genres,
while a reward function can be specifically designed 
for a single task.
We leave the learning of a generic summarisation evaluation metric 
for future work.
%

%
%

\section{Summary-Level Correlation Study}
\label{sec:motivation}
In this section, we study the summary-level 
correlation between multiple widely used summary evaluation
metrics and human judgement scores, so as to
further motivate the need for a stronger reward 
for RL. Metrics we consider include 
ROUGE (full length F-score), BLEU \cite{DBLP:conf/acl/PapineniRWZ02} and
METEOR \cite{DBLP:journals/mt/LavieD09}.
Furthermore, in line with \citet{chaganty-mussmann-liang:2018:Long},
we also use the cosine similarity between
the embeddings of the generated summary and the
reference summary as metrics: we use 
InferSent \cite{DBLP:conf/corr/ConneauKSBB17} and
BERT-Large-Cased \cite{DBLP:journals/corr/abs-1810-04805} to
generate the embeddings.

The human judgement scores we use are from
\citet{chaganty-mussmann-liang:2018:Long}, collected
as follows. First, 500 news articles were randomly sampled
from the CNN/DailyMail dataset. For each news article,
four summarisation systems were used
to generate summaries: 
the \emph{seq2seq} and \emph{pointer} models proposed
by \citet{DBLP:conf/acl/SeeLM17}, and the \emph{ml}
and \emph{ml+rl} models by \citet{DBLP:conf/iclr/PaulusXS18}. 
Hence, together with the human-written reference
summaries provided in the CNN/DailyMail dataset,
each news article has five summaries.
Crowd-sourcing workers were recruited to
rate the summaries in terms of their fluency,
redundancy level and overall quality, on a 3-point Likert scale
from $-1$ to $1$. Higher scores mean better quality.
Each summary was rated by five independent workers.
We use the averaged overall quality score
for each summary as the ground-truth human judgement.

RL-based summarisation systems assume that
the summaries ranked highly by the reward function (e.g.\ ROUGE) 
are indeed ``good'' in terms of human judgement.
We define \emph{good} summaries as follows:
a summary $y$ for a news article $x$ is 
good if (\textbf{i}) the average human judgement score for $y$
is $\geq 0.5$, and (\textbf{ii}) among the five summaries for
$x$, $y$ is ranked within the top two.
To study to what extent the above assumption is true,
we not only measure the summary-level 
correlation (Spearman's $\rho$ and Pearson's $r$) 
between the reward function and human judgements,
but also count how many good summaries identified 
by the reward function are indeed good (\emph{G-Pre}), 
and how many indeed good summaries are identified
by the reward function (\emph{G-Rec}).
We normalise the reward scores and the human judgements 
to the same range. 

\begin{table}[t]
    \centering
    \small
    \begin{tabular}{l|c c c c}
    \toprule 
    Metric & $\rho$ & $r$ & G-Pre & G-Rec \\
    \midrule 
    ROUGE-1 & .290 & .304 & .392 & .428 \\ 
    ROUGE-2 & .259 & .278 & .408 & .444 \\ 
    ROUGE-L & .274 & .297 & .390 & .426 \\ 
    ROUGE-SU4 & .282 & .279 & .404 & .440 \\
    BLEU-1 & .256 & .281 & .409 & .448 \\
    BLEU-2 & .301 & .312 & .411 & .446 \\
    BLEU-3 & .317 & .312 & .409 & .444 \\
    BLEU-4 & .311 & .307 & .409 & .446 \\
    BLEU-5 & .308 & .303 & .420 & .459 \\
    METEOR & .305 & .285 & .409 & .444 \\
    InferSent-Cosine & \textbf{.329} & \textbf{.339} &.417 & .460\\
    BERT-Cosine & .312  & .335 & \textbf{.440} & \textbf{.484} \\
    \bottomrule
    \end{tabular}
    \caption{Quality of reward metrics. 
    G-Pre and G-Rec are the precision and recall
    rate of the ``good'' summaries identified by
    the metrics, resp.
    All metrics here require reference summaries.
    We perform stemming and stop words removal as preprosessing,
    as they help increase the correlation.
    For InferSent, 
    the embeddings of the  reference/system summaries are obtained
    by averaging the embeddings of the sentences therein.
    }
    \label{table:correlation_table}
\end{table}

From Table~\ref{table:correlation_table}, we find that
all metrics we consider have low correlation with
the human judgement. More importantly, 
their G-Pre and G-Rec scores are all below .50, which means that
more than half of the good summaries 
identified by the metrics are actually not good, 
and more than 50\% of the indeed good summaries cannot be identified by the considered metrics.
Hence, hill-climbing on these metrics can hardly guide the RL agents
to generate genuinely high-quality summaries.
These observations clearly motivate the need for better rewards.
Next, we formally formulate the reward learning task.

%

\section{Problem Formulation}
\label{sec:formulation}
We focus on reward learning for single-document summarisation 
in this work, formulated as follows.
Let $\mathcal{X}$ be the set of all input documents.
For $x \in \mathcal{X}$, let $Y_x$ be the set of all summaries
for $x$ that meet the length requirement.
A \emph{reward function} 
$R(x,y;\theta)$ measures the quality of summary 
$y$ for document $x$, where $\theta$ stands for all parameters for $R$.
Note that human judgements 
can be viewed as the ground-truth reward function,
which we denote as $R^*$ henceforth. 
At training time, suppose we have access to 
$\overline{\mathcal{X}} \subseteq \mathcal{X}$
documents and $N$ summaries for each 
$x \in \overline{\mathcal{X}}$, denoted by 
$\overline{Y}_x = \{y_{x}^1, \cdots, y_x^N\} \subseteq Y_x$.
Hence, we have $|\overline{\mathcal{X}}| \times N$
summaries at training time, and our training set
includes the $R^*$ scores for these summaries:
$\bigcup_{x\in\overline{\mathcal{X}}} \{R^*(x,y_x^1),
\cdots, R^*(x,y_x^N)\}$.
Our target is to learn a reward function $R$
that is as ``close'' to $R^*$ as possible.
Depending on the definition of ``close'', we explore
two loss functions for reward learning, detailed below.

\paragraph{Regression loss.}
We first consider reward learning as a regression problem,
by measuring the ``closeness'' between $R$ and $R^*$ by
their mean squared errors:
\begin{align}
& \mathcal{L}^{MSE}(\theta) =  \nonumber \\ 
& \frac{1}{|\overline{\mathcal{X}}|\cdot N} 
\sum_{x \in \overline{\mathcal{X}}} \sum_{i=1}^N 
[R^*(x,y_x^i)-R(x,y_x^i;\theta)]^2.
\label{eq:regression_loss}
\end{align}

\paragraph{Cross-entropy loss.}
An alternative definition of ``closeness'' is to 
measure the ``agreement'' between $R$ and $R^*$,
i.e.\ for a pair of summaries, whether $R$ and $R^*$
prefer the same summary. 
For two summaries $y_x^{i}, y_x^{j} \in \overline{Y}_x$,
we estimate the likelihood that $R$ prefers
$y_x^{i}$ over $y_x^{j}$ as 
\begin{align}
P(y_x^{i} \succ y_x^{j}) & = \frac{\exp(r^i)}{
\exp(r^i)+ \exp(r^j)},
\label{eq:pref_loss}
\end{align}
where $r^i  = R(x,y_x^{i};\theta)$, 
$r^j = R(x,y_x^{j};\theta)$.
Note that for each $x \in \overline{\mathcal{X}}$,
we have $N$ summaries available in $\overline{Y}_{x}$.
Hence we can construct $N\cdot(N-1)/2$ pairs of summaries
for each input $x$.
Our target is to minimise the ``disagreement''
between $R^*$ and $R$ on the $|\overline{\mathcal{X}}|\cdot
N\cdot(N-1)/2$ pairs of summaries, formally defined
as the cross-entropy loss below:
\begin{align}
& \mathcal{L}^{CE}(\theta)  = 
- \frac{1}{|\overline{\mathcal{X}}|N(N-1)/2} 
\sum_{x \in \overline{\mathcal{X}}} 
\sum_{i=1}^N \sum_{j>i}^N 
\{ \nonumber \\
& \mathbbm{1}[R^*(x,y_x^i)>
R^*(x,y_x^j)] \log P(y_x^i \succ y_x^j) + 
\nonumber \\
& \mathbbm{1}[R^*(x,y_x^j)>
R^*(x,y_x^i)]\log P(y_x^j \succ y_x^i) \},
\label{eq:ce_loss}
\end{align}
where $\mathbbm{1}$ is the indicator function.
%
%
Next, we will introduce our reward learning
model that minimises the losses
defined in Eq.~\eqref{eq:regression_loss}
and \eqref{eq:ce_loss}.

\section{Reward Learning Model}
\label{sec:model}

We explore two neural architectures for $R(x,y;\theta)$:
\emph{Multi-Layer Perceptron (MLP)}  
and \emph{Similarity-Redundancy Matrix (SimRed)}, detailed
below. 

\subsection{MLP}
\label{subsec:mlp}
A straightforward method for learning $R(x,y_x;\theta)$ 
is to encode the input document $x$ and summary $y$ as two embeddings,
and feed the concatenated embedding into a single-layer
MLP to minimise
the loss functions
Eq.~\eqref{eq:regression_loss} and \eqref{eq:ce_loss}.
We consider three text encoders to vectorise $x$ and $y_x$. 
In supplementary material, we
provide figures to
further illustrate the architectures of these text encoders.

%

\paragraph{CNN-RNN.} 
We use convolutional neural networks (CNNs) to encode
the sentences in the input text, and feed the sentence embeddings
into an LSTM to generate the embedding of the whole input text.
In the CNN part, convolutions with different filter 
widths are applied independently as in \cite{DBLP:journals/corr/Kim14f}.
The most relevant features are selected 
afterwards with max-over-time pooling.
In line with \citet{DBLP:conf/naacl/NarayanCL18}, we 
reverse the order of sentence embeddings before 
feeding them into the LSTM. 
This encoder network yields strong performance
on summarisation and sentence classification tasks
\cite{DBLP:conf/acl/YuLCNCPC18,DBLP:conf/naacl/NarayanCL18}.

\paragraph{PMeans-RNN.}
%
PMeans is a simple yet powerful sentence encoding
method \cite{DBLP:journals/corr/abs-1803-01400}.
PMeans encodes a sentence by computing the \emph{power means}
of the embeddings of the words in the sentence.
PMeans uses a parameter $p$ to control the weights
for each word embedding: 
with $p=1$, each word element is weighted 
equally, and  
with the increase of $p$, it assigns higher weights
to the elements with higher values. With $p=+\infty$,
PMeans is equivalent to element-wise max-pooling.
The output of PMeans 
is passed to an LSTM to produce the final document embedding. 
Note that only the LSTM is trainable; the $p$ value
is decided by the system designer. 



\paragraph{BERT.}
We use the pre-trained BERT-Large-Cased 
model to encode news articles and summaries. 
The hidden state of the final layer that corresponds to the first token (i.e. ``[CLS]") is taken as embedding.
Note that the pre-trained BERT models can only
encode texts with at most 512 tokens. In line with
\citet{DBLP:journals/corr/abs-1901-08634},
we therefore use a sliding window approach with 
the offset size of 128 tokens
to encode overlength summaries and news articles.
We do not fine-tune the BERT model because our dataset is relatively small (only 2,500 summaries
and 500 news articles), and the sliding-window 
of BERT
requires
much resources to fine-tune.

\subsection{Similarity-Redundancy Matrix (SimRed)}
\label{subsec:simred}
Good summaries should be more \emph{informative} (i.e.\
\OS[include more important information 
of the input documents]{contain information of higher importance from the input documents}) and less \emph{redundant} than bad summaries. 
Based on this intuition, we propose the SimRed architecture, 
which explicitly measures the informativeness and redundancy
of summary $y_x$ for document $x$.
%
%
SimRed maintains a \emph{Similarity} matrix
and a 
\emph{Redundancy} matrix. 
In the Similarity matrix,
cell $(i,j)$ is the cosine similarity between the
embeddings of the $i$th sentence in summary $y_x$
and the $j$th sentence in document $x$.
In the Redundancy matrix, each cell contains the
square of the cosine similarity of a pair of 
sentences in summary $y_x$.
%
We use the average over the Similarity matrix cells 
to measure the informativeness of $y_x$, the average over the 
Redundancy matrix cells to measure the redundancy,
and compute the weighted sum of these two averaged values
to yield the reward $R$:
\begin{align}
    R_{\operatorname{SimRed}}(y_x,x) & = \frac{\alpha}{NM} \sum_{i=1}^N \sum_{j=1}^M 
    \cos(s_i,d_j) \nonumber \\
    - \frac{1 - \alpha}{N(N-1)/2} & \sum_{k=1}^N\sum_{l>k}^N (
    \cos(s_k, s_l))^2,
    \label{eq:simred}
\end{align}

where $s_i, i=1,\cdots,N$ indicates the embedding of the 
$i$th sentence in summary $y_x$, and $d_j, j=1,\cdots,M$
indicates the embedding of the $j$th sentence in document $x$.
%
The sentence embeddings are generated using
CNN, PMeans and BERT as described in \S\ref{subsec:mlp}. 
Because PMeans does not have trainable parameters 
and BERT is kept fixed, 
we put a trainable layer on top of them. 
\section{Reward Quality Evaluation}
\label{sec:reward_quality}

\paragraph{Experimental Setup.}
We perform 5-fold cross-validation on the 2,500 human
summaries (described in \S\ref{sec:motivation}) 
to measure the performance of our reward $R$. 
In each fold, the data is split with 
ratio 64:16:20 for training, validation and test.

The parameters of our model are detailed as follows,
decided in a pilot study on one fold of the data split.
The CNN-RNN encoder (see \S\ref{subsec:mlp})
uses filter widths 1-10 for the CNN part, 
and uses a unidirectional LSTM with a single 
layer whose dimension is 600 for the RNN part.
For PMeans, we obtain sentence 
embeddings for each $p \in \left\{-\infty, +\infty, 1, 2\right\}$ 
and concatenate them per sentence. 
Both these two encoders use the pre-trained GloVe word 
embeddings \cite{pennington-etal-2014-glove}. 
On top of these encoders, we use an MLP with 
one hidden ReLU layer and a linear activation 
at the output layer. 
For the MLP that uses CNN-RNN and PMeans-RNN,
the dimension of its hidden layer is 100, while
for the MLP with BERT as encoder, 
the dimension of the hidden layer is 1024.
As for SimRed,
we set $\alpha$ 
(see Eq.~\eqref{eq:simred}) to be 0.85.
\FB{The trainable layer on top of BERT and PMeans -- when used with SimRed -- is a single layer perceptron whose output dimension is equal to the input dimension.}

\paragraph{Reward Quality.}
\begin{table*}[h]
    \centering
    \small
    \begin{tabular}{l l | c c c c | c c c c}
    \toprule 
    & & \multicolumn{4}{c}{\textit{Reg. loss (Eq. \eqref{eq:regression_loss})}}  & 
    \multicolumn{4}{|c}{\textit{Pref. loss (Eq. \eqref{eq:ce_loss})}} \\
    Model & Encoder & $\rho$ & $r$ & G-Pre & G-Rec 
    & $\rho$ & $r$ & G-Pre & G-Rec\\
    \midrule
    \multirow{3}{*}{MLP} &
    CNN-RNN      & .311 & .340 & .486 & .532 
                 & .318 & .335 & .481 & .524 \\
    & PMeans-RNN & .313 & .331 & .489 & .536 
                 & .354 & .375 & .502 & .556 \\
    & BERT & \textbf{.487}  & \textbf{.526} 
    & \textbf{.544} & \textbf{.597}
    & \textbf{.505} & \textbf{.531} & 
    \textbf{.556} & \textbf{.608} \\
    \midrule
    \multirow{3}{*}{SimRed} &
    CNN       & .340 & .392 & .470 & .515 
              & .396 & .443 & .499 & .549  \\
    & PMeans  & .354 & .393 & .493 & .541 
              & .370 & .374 & .507 & .551 \\
    & BERT    & .266 & .296 & .458 & .495 
              & .325 & .338 & .485 & .533 \\
    \midrule
    \multicolumn{2}{l|}{\cite{DBLP:conf/naacl/PeyrardG18}} & 
    .177 & .189 & .271 & .306 & .175 & .186 & .268 & .174 \\
    \bottomrule
    \end{tabular}
    \caption{Summary-level correlation of learned reward functions.
    All results are averaged over 5-fold cross validations. 
    Unlike the metrics in
    Table~\ref{table:correlation_table},
    all rewards in this table do not require reference summaries.}
    \label{tab:correlation}
\end{table*}

\begin{figure*}[h]
    \centering
    \includegraphics[width=.99\textwidth]{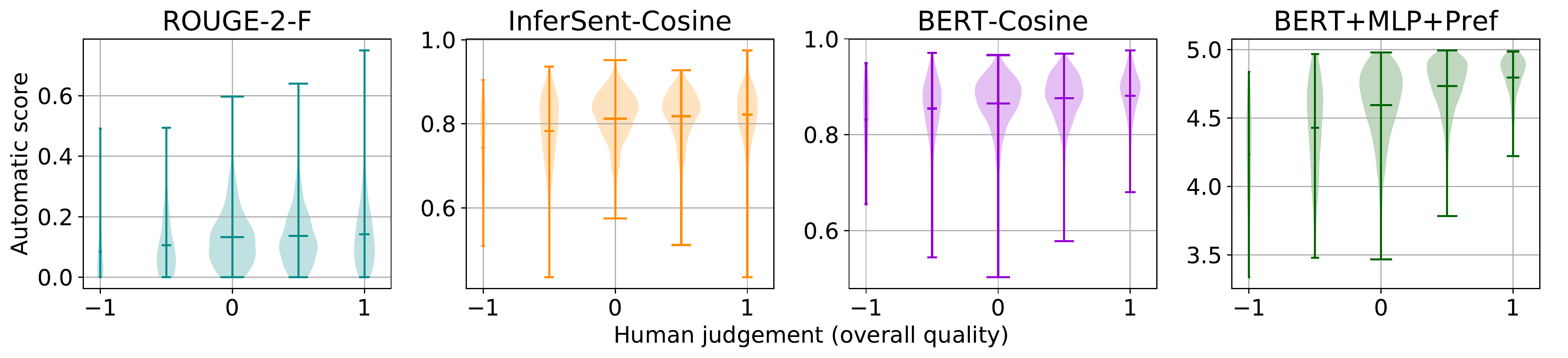}
    \caption{Distributions of some metrics/rewards for summaries
    with different human ratings. Among the four presented metrics/rewards, only BERT+MLP+Pref
    (the rightmost sub-figure) does not use reference summaries.}
    \label{fig:violin_plots}
\end{figure*}

Table~\ref{tab:correlation}
shows the quality of different reward learning models.
As a baseline, we also consider the feature-rich reward
learning method proposed by \citet{DBLP:conf/naacl/PeyrardG18}
(see \S\ref{sec:related_work}).
MLP with BERT as encoder has the best overall performance.
%
Specifically, 
BERT+MLP+Pref significantly outperforms ($p<0.05$) 
all the other models that do not use BERT+MLP, as well as the
metrics that rely on reference summaries 
(see Table~\ref{table:correlation_table}).
P-values between each pair of metrics/rewards can 
be found in the supplementary material.
%
In general, 
preference loss (Eq.~\eqref{eq:pref_loss}) yields better
performance than regression loss (Eq.~\eqref{eq:regression_loss}),
because it  
``pushes'' the reward function to provide correct 
preferences over summaries, which leads to more
precise ranking.
%


Fig.~\ref{fig:violin_plots} illustrates the 
distribution of some rewards/metrics
for summaries with different human ratings. 
In the left-most sub-figure in Fig.~\ref{fig:violin_plots},
we find that, for summaries with average human rating $1.0$ 
(the highest human rating; see \S\ref{sec:motivation}),
their average ROUGE-2 scores are similar
to those with lower human ratings, which
indicates that ROUGE-2 can hardly distinguish
the highest-quality summaries from the rest. 
%
We make similar observations for InferSent-Cosine and BERT-Cosine.
BERT+MLP+Pref provides higher scores to 
summaries with higher human ratings (the rightmost sub-figure),
although it does not use reference summaries.
This explains the strong G-Pre and G-Rec scores of BERT+MLP+Pref.
The distributions of the other metrics/rewards 
can be found in the supplementary material.
Next, we use the reward learned by BERT+MLP+Pref to
train some RL-based summarisation systems.

\section{RL-based Summarisation with Learned Rewards}
\label{sec:summary_evaluation}

We consider two RL-based summarisation systems, 
an extractive system \emph{NeuralTD} \cite{irj2019},  
and an abstractive system \emph{ExtAbsRL}
\cite{DBLP:conf/acl/BansalC18}.
%
Note that ExtAbsRL is a cross-input RL while NeuralTD is an
input-specific RL (see \S\ref{sec:related_work}).
Our study is performed on the test set
of the non-anonymised CNN/DailyMail dataset \cite{DBLP:conf/acl/SeeLM17},
which includes 11,490 news articles and one reference
summary for each article.


\subsection{RL-based Summarisation Systems}
\paragraph{NeuralTD.}
NeuralTD is an improved version of the RL-based extractive 
multi-document 
summarisation system proposed by \citet{ryang2012emnlp}.
We briefly recap the original system below.
Suppose the RL agent has selected some sentences and
has built a draft summary $d$ using the selected sentences.
The RL agent maintains a function 
$V\colon D \rightarrow \mathbb{R}$, where $D$ denotes the 
set of all possible draft summaries. 
$V(d;w)$ estimates the quality and potential of $d$, 
where $w$ are the learnable parameters in $V$.
To select which sentence to add to 
$d$ next,
the agent samples sentences $s \in S$ with distribution 
\begin{align*}
\pi(s;w) = \frac{\exp[V((d,s);w)]}{\sum_{s'\in S}\exp[V((d,s');w)]},
\end{align*}
where $S$ is the set of all sentences in the input document
that has not been added to $d$ yet, 
and $(d,s)$ is the resulting summary of concatenating $d$ and $s$.
%
%
The original system proposed by \citet{ryang2012emnlp}
models $V$ as a linear function
(i.e. $V(d;w) = w \cdot \phi(d)$, where $\phi(d)$ is the vector
for draft summary $d$). 
NeuralTD instead uses a neural network with multiple
hidden layers to approximate $V$.
\citet{irj2019} show that NeuralTD significantly
outperforms the original linear algorithm
in multiple benchmark datasets.

In line with \cite{irj2019}, 
we use the \emph{delayed rewards} in NeuralTD:
a non-zero reward is provided to the agent only
when the agent finishes the sentence selection
process (i.e.\ when agent performs the ``end-of-construction''
action). The assumption underlying this reward scheme
is that the reward function can only precisely
measure the quality of the summary when the 
summary is complete.
Besides our learned reward, in order to encourage the 
agent to select the ``lead'' sentences 
(i.e. the first three sentences in each news article),
we provide the agent with a small extra reward (0.5) 
for each ``lead'' sentence the agent chooses to extract.
The value for the extra reward (0.5) is decided in a pilot study,
in which we manually check the quality of the generated summaries
with different extra rewards (0.1, 0.3, $\cdots$, 0.9).

\paragraph{ExtAbsRL} 
has an \emph{extractor} to extract salient sentences and 
an \emph{abstractor} to 
rephrase the extracted sentences to generate abstractive summaries.
%
The abstractor is a simple encoder-aligner-decoder
model with copying mechanism, which is pre-trained using
standard supervised cross-entropy training.
The extractor, on the other hand, applies an actor-critic
RL algorithm on top of a pointer network. 
Unlike NeuralTD that uses delayed rewards,
ExtAbsRL receives a non-zero reward after adding each new sentence,
by computing the ROUGE-L score between
the 
newly 
added sentence and the corresponding 
sentence in the reference summary.
When the generated summary has more sentences than
the reference summary, 0 is given as the reward 
for the extra sentences. At the final step,
ROUGE-1 of the whole generated summary is granted as reward.

We follow the step-wise reward scheme in original ExtAbsRL
but, instead of using ROUGE-L to compute the step-wise rewards,
we apply our learned reward function to compute 
the score for the summary with and without the new 
sentence, and use their difference as reward.
Similarly, for the final step reward, we also use our learned
reward function.
In addition, we force our summariser to stop 
adding new sentences when the 
number of tokens in the generated
summary is 1.2 times as many as in the reference summary,
because we find the abstractive summaries
generated by the original ExtAbsRL algorithm 
are approximately of this length. 
%
Since the abstractor in ExtAbsRL is not RL-based,
the different reward only influences 
the extractor.
%
%
%

\subsection{Extractive Summarisation}
\label{subsec:neural_td}

Table~\ref{table:rouge_ext} presents the ROUGE scores
of our system (NeuralTD+LearnedRewards) 
and multiple state-of-the-art systems. 
The summaries generated by our system 
receive decent ROUGE metrics, but are lower
than most of the recent systems,
because our learned reward is optimised towards
high correlation with human judgement instead of ROUGE
metrics.

\begin{table}[]
    \centering
    \small
    \begin{tabular}{l l |l l l}
    \toprule 
    System & Reward & R-1 & R-2 & R-L \\ 
    \midrule 
    \cite{DBLP:conf/emnlp/KryscinskiPXS18} & R-L & 
    40.2 & 17.4 & 37.5 \\
    \cite{DBLP:conf/naacl/NarayanCL18} & R-1,2,L & 40.0 & 18.2 & 36.6 \\
    \cite{DBLP:conf/acl/BansalC18} & R-L & 41.5 & 18.7 & 37.8 \\ 
    \cite{DBLP:conf/emnlp/DongSCHC18} & R-1,2,L &41.5 & 18.7 & 37.6 \\
    \midrule
    \cite{DBLP:conf/emnlp/ZhangLWZ18} & NA &41.1 & 18.8 & 37.5 \\
    \cite{DBLP:conf/acl/ZhaoZWYHZ18} & NA & 41.6 & 19.0 & 38.0 \\
    \cite{DBLP:conf/emnlp/KedzieMD18} & NA & 39.1 & 17.9 & 35.9 \\
    \midrule 
    (ours) NeuralTD & Learned & 39.6 & 18.1 & 36.5 \\
    \bottomrule
    \end{tabular}
    \caption{Full-length ROUGE F-scores of some recent 
    RL-based (upper) and supervised (middle) extractive
    summarisation systems, as well as our system
    with learned rewards (bottom). 
     R-1/2/L stands for ROUGE-1/2/L.
    Our system maximises the 
    learned reward instead of ROUGE, hence receives lower
    ROUGE scores.
    }
    \label{table:rouge_ext}
\end{table}

To measure the human ratings on the generated summaries,
we invited five users to read and rate
the summaries from three systems: NeuralTD+LearnedReward,
the Refresh system \cite{DBLP:conf/naacl/NarayanCL18}
and the extractive version of the ExtAbsRL system, which
only extracts salient sentences and does not
apply sentence rephrasing.
%
We selected Refresh and ExtAbsRL 
because they both have been
reported to receive higher human ratings than
the strong system proposed by \citet{DBLP:conf/acl/SeeLM17}.
%

We randomly sampled 30 news articles from
the test set in CNN/DailyMail,
and asked the five participants to rate the three summaries
for each article on a 3-point Likert scale 
from 1 to 3, where
higher scores mean better overall quality.
We asked them to consider the \emph{informativeness}
(whether the summary contains most important information
in the article) and \emph{conciseness} (whether the summary is
concise and does not contain redundant information)
in their ratings.

\begin{table}[]
    \centering
    \small
    \begin{tabular}{r | c c c}
    \toprule 
     & Ours & Refresh & ExtAbsRL \\ 
     \midrule
     Avg. Human Rating & \textbf{2.52} & 2.27 & 1.66 \\
     Best\% & \textbf{70.0} & 33.3 & 6.7 \\
     \bottomrule
    \end{tabular}
    \caption{Human evaluation on extractive summaries. Our system
    receives significantly higher human ratings on average. 
    ``Best\%'': in how many percentage of documents 
    a system receives the highest human rating.}
    \label{table:human_ext}
\end{table}

Table~\ref{table:human_ext} presents the human evaluation results.
summaries generated by NeuralTD receives significantly higher human evaluation
scores than those by Refresh ($p=0.0088$, double-tailed t-test) 
and ExtAbsRL ($p \ll 0.01$).
Also, the average human rating for Refresh is significantly
higher ($p \ll 0.01$) than ExtAbsRL, 
despite 
receiving significantly 
higher ROUGE scores
than both Refresh and NeuralTD (see Table~\ref{table:rouge_ext}).
We find that the summaries generated by Ext\-Abs\-RL
include more tokens (94.5) than those generated
by Refresh (83.4) and NeuralTD (85.6).
\citet{sun2019naacl} recently show 
that, for summaries whose lengths are 
in the range of 50 to 110 tokens, 
longer summaries receive higher ROUGE-F1 scores. 
We believe this is the reason why ExtAbsRL
has higher ROUGE scores. 
On the other hand, ExtAbsRL extracts more redundant sentences:
four out of 30 summaries by Ext\-Abs\-RL include redundant
sentences, while Refresh and NeuralTD do not 
generate summaries with two identical sentences therein.
Users are sensitive to the redundant sentences in summaries:
the average human rating for redundant summaries
is 
1.2, lower than the average rating for
the other summaries generated by ExtAbsRL (1.66).
To summarise, by using our learned reward function
in training an extractive RL summariser (NeuralTD),
the generated summaries receive significantly higher
human ratings than the state-of-the-art systems.

\subsection{Abstractive Summarisation}
\label{subsec:ext_abs_rl}

\begin{table}[]
    \centering
    \small
    \begin{tabular}{ l | l l l | c c}
    \toprule
    Reward & R-1 & R-2 & R-L & Human & Pref\% \\
    \midrule
    R-L (original) & 40.9 & 17.8 & 38.5 & 1.75 & 15 \\
    Learned (ours) & 39.2 & 17.4 & 37.5 & \textbf{2.20} & \textbf{75} \\
    \bottomrule
    \end{tabular}
    \caption{Performance of ExtAbsRL with different reward functions,
    measured in terms of ROUGE (center) and 
    human judgements (right).
    Using our learned reward 
    yields significantly
    ($p=0.0057$) higher average human rating. 
    ``Pref\%'': in how many percentage of documents 
    a system receives the higher human rating.
    }
    \label{table:rouge_abs}
\end{table}

Table~\ref{table:rouge_abs} compares the ROUGE scores
of using different rewards to train the
extractor in ExtAbsRL (the abstractor is pre-trained,
and is applied to rephrase the extracted sentences).
Again, when ROUGE is used as rewards, the generated
summaries have higher ROUGE scores.

We randomly sampled 20 documents from
the test set in CNN/DailyMail and asked
three users to rate the quality of the two summaries
generated with different rewards.
We asked the users to consider not only the
informativeness and conciseness of summaries,
but also their grammaticality and faithfulness (whether
the information in the summary is consistent with that
in the original news).
%
It is clear from Table~\ref{table:rouge_abs} 
that using the learned reward helps the RL-based system 
generate summaries with significantly higher human ratings.
%
Furthermore, we note that the overall human ratings for 
the abstractive summaries are lower than the extractive
summaries (compared to Table~\ref{table:human_ext}).
%
%
Qualitative analysis suggests that the poor overall 
rating may be caused by occasional 
information inconsistencies between a 
summary and its source text: for instance, a sentence in the
source article reads 
\textit{``after Mayweather was almost two hours late for his workout , Pacquiao has promised to be on time''},
but the generated summary outputs 
\textit{``Mayweather has promised to be on time for the fight''}.
High redundancy is another reason for the low human ratings:
ExtAbsRL generates six summaries with redundant sentences 
when applying ROUGE-L as reward, 
while the number drops to two when the learned reward is applied.
%

\section{Conclusion \& Discussion}
\label{sec:conclusion}

In this work, we focus on Reinforcement Learning (RL)
based summarisation,
and 
propose a reward function
directly learned from human ratings on summaries' overall quality. 
%
Our reward function only takes the source text
and the generated summary as input (i.e. does not require reference
summaries), 
and correlates significantly
better with human judgements than existing metrics
(e.g. ROUGE and METEOR, which require reference summaries).
We use our learned reward to train both extractive and abstractive
summarisation systems.
Experiments show that the summaries
generated from our learned reward 
outperform those by the state-of-the-art systems,
in terms of human judgements.
Considering that our reward is learned from only 2,500 human ratings 
on 500 summaries, while the state-of-the-art systems
require two orders of magnitude (287k) more 
document-reference pairs for training,
this work clearly shows that reward learning plus RL-based
summarisation is able to leverage a relatively
small set of human rating scores to produce high-quality summaries.
%

For future work, we plan to test our method in other summarisation
tasks (e.g. multi-document summarisation) 
and downstream tasks of summarisation
(e.g. investigating whether users can correctly answer
questions by reading our summaries 
instead of the original documents).
Also, we believe the ``learning reward from human judgements''
idea has potential to boost the performance of RL
in other natural language generation applications,
e.g. translation, sentence simplification and dialogue generation.

\section*{Acknowledgements}
This work has been supported by the German
Research Foundation (DFG), as part of the 
QA-EduInf project (
GU 798/18-1 and 
RI 803/12-1) and through the 
German-Israeli Project Cooperation 
(DIP, 
DA 1600/1-1 and 
GU 798/17-1).

\bibliography{references}

\begin{thebibliography}{41}
\expandafter\ifx\csname natexlab\endcsname\relax\def\natexlab#1{#1}\fi

\bibitem[{{Alberti} et~al.(2019){Alberti}, {Lee}, and
  {Collins}}]{DBLP:journals/corr/abs-1901-08634}
Chris {Alberti}, Kenton {Lee}, and Michael {Collins}. 2019.
\newblock \href {http://arxiv.org/abs/1901.08634} {{A BERT Baseline for the
  Natural Questions}}.
\newblock \emph{arXiv e-prints}.

\bibitem[{Arumae and Liu(2019)}]{DBLP:journals/corr/abs-1904-02321}
Kristjan Arumae and Fei Liu. 2019.
\newblock \href {https://doi.org/http://dx.doi.org/10.18653/v1/N19-1264}
  {Guiding extractive summarization with question-answering rewards}.
\newblock In \emph{Proceedings of the 2019 Conference of the North American
  Chapter of the Association for Computational Linguistics: Human Language
  Technologies, Volume 1 (Long Papers)}, pages 2566--2577, Minneapolis, USA.

\bibitem[{Bowman et~al.(2015)Bowman, Angeli, Potts, and
  Manning}]{DBLP:conf/emnlp/BowmanAPM15}
Samuel~R. Bowman, Gabor Angeli, Christopher Potts, and Christopher~D. Manning.
  2015.
\newblock \href {http://aclweb.org/anthology/D/D15/D15-1075.pdf} {A large
  annotated corpus for learning natural language inference}.
\newblock In \emph{Proceedings of the 2015 Conference on Empirical Methods in
  Natural Language Processing}, pages 632--642, Lisbon, Portugal.

\bibitem[{Chaganty et~al.(2018)Chaganty, Mussmann, and
  Liang}]{chaganty-mussmann-liang:2018:Long}
Arun Chaganty, Stephen Mussmann, and Percy Liang. 2018.
\newblock \href {http://www.aclweb.org/anthology/P18-1060} {{The price of
  debiasing automatic metrics in natural language evalaution}}.
\newblock In \emph{Proceedings of the 56th Annual Meeting of the Association
  for Computational Linguistics, Volume 1: Long Papers}, pages 643--653,
  Melbourne, Australia.

\bibitem[{Chen and Bansal(2018)}]{DBLP:conf/acl/BansalC18}
Yen{-}Chun Chen and Mohit Bansal. 2018.
\newblock \href {https://aclanthology.info/papers/P18-1063/p18-1063} {Fast
  abstractive summarization with reinforce-selected sentence rewriting}.
\newblock In \emph{Proceedings of the 56th Annual Meeting of the Association
  for Computational Linguistics, Volume 1: Long Papers}, pages 675--686,
  Melbourne, Australia.

\bibitem[{Conneau et~al.(2017)Conneau, Kiela, Schwenk, Barrault, and
  Bordes}]{DBLP:conf/corr/ConneauKSBB17}
Alexis Conneau, Douwe Kiela, Holger Schwenk, Lo{\"{\i}}c Barrault, and Antoine
  Bordes. 2017.
\newblock \href {https://aclanthology.info/papers/D17-1070/d17-1070}
  {Supervised learning of universal sentence representations from natural
  language inference data}.
\newblock In \emph{Proceedings of the 2017 Conference on Empirical Methods in
  Natural Language Processing}, pages 670--680, Copenhagen, Denmark.

\bibitem[{Devlin et~al.(2019)Devlin, Chang, Lee, and
  Toutanova}]{DBLP:journals/corr/abs-1810-04805}
Jacob Devlin, Ming-Wei Chang, Kenton Lee, and Kristina Toutanova. 2019.
\newblock \href {https://doi.org/10.18653/v1/N19-1423} {{BERT}: Pre-training of
  deep bidirectional transformers for language understanding}.
\newblock In \emph{Proceedings of the 2019 Conference of the North {A}merican
  Chapter of the Association for Computational Linguistics: Human Language
  Technologies, Volume 1 (Long and Short Papers)}, pages 4171--4186,
  Minneapolis, Minnesota.

\bibitem[{Dong et~al.(2018)Dong, Shen, Crawford, van Hoof, and
  Cheung}]{DBLP:conf/emnlp/DongSCHC18}
Yue Dong, Yikang Shen, Eric Crawford, Herke van Hoof, and Jackie Chi~Kit
  Cheung. 2018.
\newblock \href {https://aclanthology.info/papers/D18-1409/d18-1409}
  {Banditsum: Extractive summarization as a contextual bandit}.
\newblock In \emph{Proceedings of the 2018 Conference on Empirical Methods in
  Natural Language Processing}, pages 3739--3748, Brussels, Belgium.

\bibitem[{Gao et~al.(2018)Gao, Meyer, and Gurevych}]{DBLP:conf/emnlp/GaoMG18}
Yang Gao, Christian~M. Meyer, and Iryna Gurevych. 2018.
\newblock \href {https://aclanthology.info/papers/D18-1445/d18-1445} {{APRIL:}
  interactively learning to summarise by combining active preference learning
  and reinforcement learning}.
\newblock In \emph{Proceedings of the 2018 Conference on Empirical Methods in
  Natural Language Processing}, pages 4120--4130, Brussels, Belgium.

\bibitem[{{Gao} et~al.(2019){Gao}, {Meyer}, and {Gurevych}}]{irj2019}
Yang {Gao}, Christian~M. {Meyer}, and Iryna {Gurevych}. 2019.
\newblock \href {http://arxiv.org/abs/1906.02923} {{Preference-based
  Interactive Multi-Document Summarisation}}.
\newblock \emph{arXiv e-prints}.

\bibitem[{Gao et~al.(2019)Gao, Meyer, Mesgar, and
  Gurevych}]{DBLP:conf/ijcai/0023MMG19}
Yang Gao, Christian~M. Meyer, Mohsen Mesgar, and Iryna Gurevych. 2019.
\newblock \href {https://doi.org/10.24963/ijcai.2019/326} {Reward learning for
  efficient reinforcement learning in extractive document summarisation}.
\newblock In \emph{Proceedings of the Twenty-Eighth International Joint
  Conference on Artificial Intelligence}, pages 2350--2356, Macao, China.

\bibitem[{Grusky et~al.(2018)Grusky, Naaman, and
  Artzi}]{DBLP:conf/naacl/GruskyNA18}
Max Grusky, Mor Naaman, and Yoav Artzi. 2018.
\newblock \href {https://aclanthology.info/papers/N18-1065/n18-1065} {Newsroom:
  {A} dataset of 1.3 million summaries with diverse extractive strategies}.
\newblock In \emph{Proceedings of the 2018 Conference of the North American
  Chapter of the Association for Computational Linguistics: Human Language
  Technologies, Volume 1 (Long Papers)}, pages 708--719, New Orleans,
  Louisiana, USA.

\bibitem[{Hermann et~al.(2015)Hermann, Kocisk{\'{y}}, Grefenstette, Espeholt,
  Kay, Suleyman, and Blunsom}]{DBLP:conf/nips/HermannKGEKSB15}
Karl~Moritz Hermann, Tom{\'{a}}s Kocisk{\'{y}}, Edward Grefenstette, Lasse
  Espeholt, Will Kay, Mustafa Suleyman, and Phil Blunsom. 2015.
\newblock \href
  {http://papers.nips.cc/paper/5945-teaching-machines-to-read-and-comprehend}
  {Teaching machines to read and comprehend}.
\newblock In \emph{Advances in Neural Information Processing Systems 28: Annual
  Conference on Neural Information Processing Systems 2015}, pages 1693--1701,
  Montreal, Quebec, Canada.

\bibitem[{Kedzie et~al.(2018)Kedzie, McKeown, and
  III}]{DBLP:conf/emnlp/KedzieMD18}
Chris Kedzie, Kathleen~R. McKeown, and Hal~Daum{\'{e}} III. 2018.
\newblock \href {https://aclanthology.info/papers/D18-1208/d18-1208} {Content
  selection in deep learning models of summarization}.
\newblock In \emph{Proceedings of the 2018 Conference on Empirical Methods in
  Natural Language Processing}, pages 1818--1828, Brussels, Belgium.

\bibitem[{Kim(2014)}]{DBLP:journals/corr/Kim14f}
Yoon Kim. 2014.
\newblock \href {http://aclweb.org/anthology/D/D14/D14-1181.pdf} {Convolutional
  neural networks for sentence classification}.
\newblock In \emph{Proceedings of the 2014 Conference on Empirical Methods in
  Natural Language Processing}, pages 1746--1751, Doha, Qatar.

\bibitem[{Kreutzer et~al.(2017)Kreutzer, Sokolov, and
  Riezler}]{DBLP:conf/acl/KreutzerSR17}
Julia Kreutzer, Artem Sokolov, and Stefan Riezler. 2017.
\newblock \href {https://doi.org/10.18653/v1/P17-1138} {Bandit structured
  prediction for neural sequence-to-sequence learning}.
\newblock In \emph{Proceedings of the 55th Annual Meeting of the Association
  for Computational Linguistics, Volume 1: Long Papers}, pages 1503--1513,
  Vancouver, Canada.

\bibitem[{Kreutzer et~al.(2018)Kreutzer, Uyheng, and Riezler}]{kreutzer2018b}
Julia Kreutzer, Joshua Uyheng, and Stefan Riezler. 2018.
\newblock Reliability and learnability of human bandit feedback for
  sequence-to-sequence reinforcement learning.
\newblock In \emph{Proceedings of the 56th Annual Meeting of the Association
  for Computational Linguistics, Volume 1: Long Papers}, pages 1777--1788,
  Melbourne, Australia.

\bibitem[{Kryscinski et~al.(2018)Kryscinski, Paulus, Xiong, and
  Socher}]{DBLP:conf/emnlp/KryscinskiPXS18}
Wojciech Kryscinski, Romain Paulus, Caiming Xiong, and Richard Socher. 2018.
\newblock \href {https://aclanthology.info/papers/D18-1207/d18-1207} {Improving
  abstraction in text summarization}.
\newblock In \emph{Proceedings of the 2018 Conference on Empirical Methods in
  Natural Language Processing}, pages 1808--1817, Brussels, Belgium.

\bibitem[{Lavie and Denkowski(2009)}]{DBLP:journals/mt/LavieD09}
Alon Lavie and Michael~J. Denkowski. 2009.
\newblock \href {https://doi.org/10.1007/s10590-009-9059-4} {The meteor metric
  for automatic evaluation of machine translation}.
\newblock \emph{Machine Translation}, 23(2-3):105--115.

\bibitem[{Lin(2004{\natexlab{a}})}]{lin2004looking}
Chin-Yew Lin. 2004{\natexlab{a}}.
\newblock Looking for a few good metrics: Rouge and its evaluation.
\newblock In \emph{NTCIR Workshop}.

\bibitem[{Lin(2004{\natexlab{b}})}]{lin2004rouge}
Chin-Yew Lin. 2004{\natexlab{b}}.
\newblock \href {http://aclweb.org/anthology/W04-1013} {{ROUGE}: {A} package
  for automatic evaluation of summaries}.
\newblock In \emph{ACL Workshop ``Text Summarization Branches Out''}.

\bibitem[{Louis and Nenkova(2013)}]{DBLP:journals/coling/LouisN13}
Annie Louis and Ani Nenkova. 2013.
\newblock \href {https://doi.org/10.1162/COLI\_a\_00123} {Automatically
  assessing machine summary content without a gold standard}.
\newblock \emph{Computational Linguistics}, 39(2):267--300.

\bibitem[{Narayan et~al.(2018{\natexlab{a}})Narayan, Cardenas,
  Papasarantopoulos, Cohen, Lapata, Yu, and Chang}]{DBLP:conf/acl/YuLCNCPC18}
Shashi Narayan, Ronald Cardenas, Nikos Papasarantopoulos, Shay~B. Cohen,
  Mirella Lapata, Jiangsheng Yu, and Yi~Chang. 2018{\natexlab{a}}.
\newblock \href {https://aclanthology.info/papers/P18-1188/p18-1188} {Document
  modeling with external attention for sentence extraction}.
\newblock In \emph{Proceedings of the 56th Annual Meeting of the Association
  for Computational Linguistics, Volume 1: Long Papers}, pages 2020--2030,
  Melbourne, Australia.

\bibitem[{Narayan et~al.(2018{\natexlab{b}})Narayan, Cohen, and
  Lapata}]{DBLP:conf/naacl/NarayanCL18}
Shashi Narayan, Shay~B. Cohen, and Mirella Lapata. 2018{\natexlab{b}}.
\newblock \href {https://aclanthology.info/papers/N18-1158/n18-1158} {Ranking
  sentences for extractive summarization with reinforcement learning}.
\newblock In \emph{Proceedings of the 2018 Conference of the North American
  Chapter of the Association for Computational Linguistics: Human Language
  Technologies, Volume 1 (Long Papers)}, pages 1747--1759, New Orleans,
  Louisiana, USA.

\bibitem[{Ng and Russell(2000)}]{DBLP:conf/icml/NgR00}
Andrew~Y. Ng and Stuart~J. Russell. 2000.
\newblock Algorithms for inverse reinforcement learning.
\newblock In \emph{Proceedings of the Seventeenth International Conference on
  Machine Learning}, pages 663--670, Stanford University, Stanford, CA, USA.

\bibitem[{Novikova et~al.(2017)Novikova, Dusek, Curry, and
  Rieser}]{DBLP:conf/emnlp/NovikovaDCR17}
Jekaterina Novikova, Ondrej Dusek, Amanda~Cercas Curry, and Verena Rieser.
  2017.
\newblock \href {https://aclanthology.info/papers/D17-1238/d17-1238} {Why we
  need new evaluation metrics for {NLG}}.
\newblock In \emph{Proceedings of the 2017 Conference on Empirical Methods in
  Natural Language Processing}, pages 2241--2252, Copenhagen, Denmark.

\bibitem[{Papineni et~al.(2002)Papineni, Roukos, Ward, and
  Zhu}]{DBLP:conf/acl/PapineniRWZ02}
Kishore Papineni, Salim Roukos, Todd Ward, and Wei{-}Jing Zhu. 2002.
\newblock \href {http://www.aclweb.org/anthology/P02-1040.pdf} {{BLEU}: a
  method for automatic evaluation of machine translation}.
\newblock In \emph{Proceedings of the 40th Annual Meeting of the Association
  for Computational Linguistics}, pages 311--318, Philadelphia, PA, USA.

\bibitem[{Pasunuru and Bansal(2018)}]{DBLP:conf/naacl/PasunuruB18}
Ramakanth Pasunuru and Mohit Bansal. 2018.
\newblock \href {https://aclanthology.info/papers/N18-2102/n18-2102}
  {Multi-reward reinforced summarization with saliency and entailment}.
\newblock In \emph{Proceedings of the 2018 Conference of the North American
  Chapter of the Association for Computational Linguistics: Human Language
  Technologies, Volume 2 (Short Papers)}, pages 646--653, New Orleans,
  Louisiana, USA.

\bibitem[{Paulus et~al.(2018)Paulus, Xiong, and
  Socher}]{DBLP:conf/iclr/PaulusXS18}
Romain Paulus, Caiming Xiong, and Richard Socher. 2018.
\newblock \href {https://openreview.net/forum?id=HkAClQgA-} {A deep reinforced
  model for abstractive summarization}.
\newblock In \emph{6th International Conference on Learning Representations,
  Conference Track Proceedings}, Vancouver, BC, Canada.

\bibitem[{Pennington et~al.(2014)Pennington, Socher, and
  Manning}]{pennington-etal-2014-glove}
Jeffrey Pennington, Richard Socher, and Christopher Manning. 2014.
\newblock \href {https://doi.org/10.3115/v1/D14-1162} {Glove: Global vectors
  for word representation}.
\newblock In \emph{Proceedings of the 2014 Conference on Empirical Methods in
  Natural Language Processing (EMNLP)}, pages 1532--1543, Doha, Qatar.

\bibitem[{Peyrard et~al.(2017)Peyrard, Botschen, and
  Gurevych}]{DBLP:conf/emnlp/PeyrardBG17}
Maxime Peyrard, Teresa Botschen, and Iryna Gurevych. 2017.
\newblock \href {https://aclanthology.info/papers/W17-4510/w17-4510} {Learning
  to score system summaries for better content selection evaluation}.
\newblock In \emph{Proceedings of the Workshop on New Frontiers in
  Summarization}, pages 74--84, Copenhagen, Denmark.

\bibitem[{Peyrard and Gurevych(2018)}]{DBLP:conf/naacl/PeyrardG18}
Maxime Peyrard and Iryna Gurevych. 2018.
\newblock \href {https://aclanthology.info/papers/N18-2103/n18-2103} {Objective
  function learning to match human judgements for optimization-based
  summarization}.
\newblock In \emph{Proceedings of the 2018 Conference of the North American
  Chapter of the Association for Computational Linguistics: Human Language
  Technologies, Volume 2 (Short Papers)}, pages 654--660, New Orleans,
  Louisiana, USA.

\bibitem[{Rajpurkar et~al.(2016)Rajpurkar, Zhang, Lopyrev, and
  Liang}]{DBLP:conf/emnlp/RajpurkarZLL16}
Pranav Rajpurkar, Jian Zhang, Konstantin Lopyrev, and Percy Liang. 2016.
\newblock \href {http://aclweb.org/anthology/D/D16/D16-1264.pdf} {{SQuAD: 100,
  000+ Questions for Machine Comprehension of Text}}.
\newblock In \emph{Proceedings of the 2016 Conference on Empirical Methods in
  Natural Language Processing}, pages 2383--2392, Austin, Texas, USA.

\bibitem[{Rioux et~al.(2014)Rioux, Hasan, and Chali}]{rioux2014emnlp}
Cody Rioux, Sadid~A. Hasan, and Yllias Chali. 2014.
\newblock \href {http://aclweb.org/anthology/D/D14/D14-1075.pdf} {Fear the
  {REAPER:} {A} system for automatic multi-document summarization with
  reinforcement learning}.
\newblock In \emph{Proceedings of the 2014 Conference on Empirical Methods in
  Natural Language Processing, {EMNLP} 2014}, pages 681--690, Doha, Qatar.

\bibitem[{{R{\"u}ckl{\'e}} et~al.(2018){R{\"u}ckl{\'e}}, {Eger}, {Peyrard}, and
  {Gurevych}}]{DBLP:journals/corr/abs-1803-01400}
Andreas {R{\"u}ckl{\'e}}, Steffen {Eger}, Maxime {Peyrard}, and Iryna
  {Gurevych}. 2018.
\newblock \href {http://arxiv.org/abs/1803.01400} {{Concatenated Power Mean
  Word Embeddings as Universal Cross-Lingual Sentence Representations}}.
\newblock \emph{arXiv e-prints}.

\bibitem[{Ryang and Abekawa(2012)}]{ryang2012emnlp}
Seonggi Ryang and Takeshi Abekawa. 2012.
\newblock \href {http://www.aclweb.org/anthology/D12-1024} {Framework of
  automatic text summarization using reinforcement learning}.
\newblock In \emph{Proceedings of the 2012 Joint Conference on Empirical
  Methods in Natural Language Processing and Computational Natural Language
  Learning}, pages 256--265, Jeju Island, Korea.

\bibitem[{See et~al.(2017)See, Liu, and Manning}]{DBLP:conf/acl/SeeLM17}
Abigail See, Peter~J. Liu, and Christopher~D. Manning. 2017.
\newblock \href {https://doi.org/10.18653/v1/P17-1099} {Get to the point:
  Summarization with pointer-generator networks}.
\newblock In \emph{Proceedings of the 55th Annual Meeting of the Association
  for Computational Linguistics, Volume 1: Long Papers}, pages 1073--1083,
  Vancouver, Canada.

\bibitem[{Sun et~al.(2019)Sun, Shapira, Dagan, and Nenkova}]{sun2019naacl}
Simeng Sun, Ori Shapira, Ido Dagan, and Ani Nenkova. 2019.
\newblock \href {https://doi.org/10.18653/v1/W19-2303} {{How to Compare
  Summarizers without Target Length? Pitfalls, Solutions and Re-Examination of
  the Neural Summarization Literature}}.
\newblock In \emph{Proceedings of the Workshop on Methods for Optimizing and
  Evaluating Neural Language Generation}, pages 21--29, Minneapolis, Minnesota.

\bibitem[{Williams et~al.(2018)Williams, Nangia, and
  Bowman}]{DBLP:conf/naacl/WilliamsNB18}
Adina Williams, Nikita Nangia, and Samuel~R. Bowman. 2018.
\newblock \href {https://aclanthology.info/papers/N18-1101/n18-1101} {A
  broad-coverage challenge corpus for sentence understanding through
  inference}.
\newblock In \emph{Proceedings of the 2018 Conference of the North American
  Chapter of the Association for Computational Linguistics: Human Language
  Technologies, Volume 1 (Long Papers)}, pages 1112--1122, New Orleans,
  Louisiana, USA.

\bibitem[{Zhang et~al.(2018)Zhang, Lapata, Wei, and
  Zhou}]{DBLP:conf/emnlp/ZhangLWZ18}
Xingxing Zhang, Mirella Lapata, Furu Wei, and Ming Zhou. 2018.
\newblock \href {https://aclanthology.info/papers/D18-1088/d18-1088} {Neural
  latent extractive document summarization}.
\newblock In \emph{Proceedings of the 2018 Conference on Empirical Methods in
  Natural Language Processing}, pages 779--784, Brussels, Belgium.

\bibitem[{Zhou et~al.(2018)Zhou, Yang, Wei, Huang, Zhou, and
  Zhao}]{DBLP:conf/acl/ZhaoZWYHZ18}
Qingyu Zhou, Nan Yang, Furu Wei, Shaohan Huang, Ming Zhou, and Tiejun Zhao.
  2018.
\newblock \href {https://aclanthology.info/papers/P18-1061/p18-1061} {Neural
  document summarization by jointly learning to score and select sentences}.
\newblock In \emph{Proceedings of the 56th Annual Meeting of the Association
  for Computational Linguistics, Volume 1: Long Papers}, pages 654--663,
  Melbourne, Australia.

\end{thebibliography}
\bibliographystyle{acl_natbib}

\end{document}